\documentclass{article}

\usepackage{arxiv}

\usepackage[utf8]{inputenc} 
\usepackage[T1]{fontenc}    
\usepackage{hyperref}       
\usepackage{url}            
\usepackage{booktabs}       
\usepackage{amsfonts}       
\usepackage{nicefrac}       
\usepackage{microtype}      
\usepackage{lipsum}
\usepackage{float}
\usepackage{multirow}
\usepackage{array}
\usepackage{tabularx}
\usepackage{amsmath}
\usepackage{siunitx}
\usepackage{graphicx} 
\usepackage{pdflscape}
\usepackage{hyperref}

\newcolumntype{C}{>{\centering\arraybackslash}X}

\title{Accelerating Evolution: Integrating PSO Principles into Real-Coded Genetic Algorithm Crossover}
\author{
 Xiaobo Jin \\
  School of Information Engineering\\
  Taizhou Vocational College of Science \& Technology\\
  \texttt{2022001632@tzvcst.edu.cn} \\
   \And
 JiaShu Tu \\
  Traditional Chinese Medicine department\\
  Taizhou First People's Hospital\\
  \texttt{tujiashu1@126.com} \\
} 
\begin{document}
\maketitle              

\begin{abstract}
This study introduces an innovative crossover operator named Particle Swarm Optimization-inspired Crossover (PSOX), which is specifically developed for real-coded genetic algorithms. Departing from conventional crossover approaches that only exchange information between individuals within the same generation, PSOX uniquely incorporates guidance from both the current global best solution and historical optimal solutions across multiple generations. This novel mechanism enables the algorithm to maintain population diversity while simultaneously accelerating convergence toward promising regions of the search space. The effectiveness of PSOX is rigorously evaluated through comprehensive experiments on 15 benchmark test functions with diverse characteristics, including unimodal, multimodal, and highly complex landscapes. Comparative analysis against five state-of-the-art crossover operators reveals that PSOX consistently delivers superior performance in terms of solution accuracy, algorithmic stability, and convergence speed, especially when combined with an appropriate mutation strategy. Furthermore, the study provides an in-depth investigation of how different mutation rates influence PSOX's performance, yielding practical guidelines for parameter tuning when addressing optimization problems with varying landscape properties. 

\end{abstract}
\section{Introduction}
In the recent years, most of real-life problems such as Vehicle Routing Problem(VRP),Job Shop Scheduling(JSS) and Maximum Clique Problem(MCP) are NP-hard problems\cite{ref1}, which means those problems can not be solved in exact polynomial-time. To enhance solution efficiency, advanced metaheuristic frameworks such as particle swarm optimization (PSO)\cite{ref2}, Cover-Relax-Search (CRS)\cite{ref3} and ant colony optimization (ACO)\cite{ref4} have been rigorously developed and empirically validated across diverse application contexts.

Genetic algorithm (GA), a widely recognized metaheuristic approach derived from Darwinian principles of natural selection and biological evolutionary mechanisms\cite{ref5}, has demonstrated significant efficacy in solving complex optimization problems across diverse domains. In the genetic algorithm, a single chromosome is used to represent a feasible solution, and a specific number of chromosomes are utilized to model and conduct the exploration of the solution space of a given problem. The fitness value serves as a crucial metric for quantitatively assessing the quality of candidate solutions. By simulating the processes in nature, such as the selection of mates by organisms, crossover and exchange of chromosomes, genetic mutation, and natural selection, the algorithm continuously optimizes the feasible solutions. The earliest genetic algorithms (GAs) adopted binary string encoding for chromosomes, but this approach suffers from issues such as low precision and high computational complexity. In contrast, real-coding encoding is more suitable for most engineering problems. GAs that utilize real-coded chromosomes are termed Real Coded GAs\cite{ref6}.

The crossover operator with the objective of evolving superior chromosomes serves as a fundamental mechanism in genetic algorithms (GAs) by facilitating information exchange between chromosomes to explore the search space\cite{ref6,ref7}. Based on the building-block hypothesis\cite{ref8}, the crossover operator is the preeminent operator in genetic algorithms (GAs). According to this hypothesis, certain key genetic components contribute to the emergence of highly fit individuals within the population. The crossover operation effectively allows individuals to merge these essential genetic components, facilitating the creation of superior offspring. Consequently, substantial research efforts have focused on devising sophisticated real - coded crossover operators to improve the performance of Real-Coded GAs.

This study introduces PSOX, an innovative crossover operator inspired from the Particle Swarm Optimization (PSO) algorithm. We will compare the genetic algorithm using this operator with five other mainstream operators. For equitable algorithm comparison, we evaluate both our proposed methods and conventional generational genetic algorithms (GAs) using 15 benchmark problems with diverse complexity levels. We comprehensively analyze and compare the proposed operator from perspectives such as the solution accuracy, stability, and convergence speed of the problems. Moreover, through combinatorial experiments with different mutation operators and mutation rates, we investigate the interaction between our operator and different mutation operator.

The paper is structured as follows: Section 2 presents a literature review on real-coded operators, Section 3 details the proposed PSOX crossover, Section 4 describes the experimental setup for all GAs, Section 5 discusses the performance comparison of various GAs, and Section 6 provides concluding remarks.

\section{Literature review}
Numerous crossover operators for real-coded GAs have been proposed in the literature, which can be broadly categorized into two types according to the differences in the information content they incorporate. One is to generate offspring by constructing linear equations of parental information, which can be described by following equation:
	\begin{equation}
		O = \sum_{i}^{n} \alpha_{i} x_{i}
	\end{equation}

	\begin{table}
		\caption{Crossover operator by linear combination}\label{tab:linear}
		\begin{tabular}{ m{5cm} m{5cm} m{5cm} }
		\hline
		Name of crossover operation &  Formula & Parameter\\
		\hline
		Logistic Crossover\\(LX)\cite{ref10} & $ O = p^{1} + (p^{2} - p^{1})L $ & $L = \mu + s \cdot \ln \left( \frac{U}{1 - U} \right)$\\
		Flat crossover\\(FX)\cite{ref11} &  $ \begin{aligned}O = \mbox{uniform}(\min(p^1, p^2),\\ \max(p^1, p^2))\end{aligned}$ & No parameter\\
		Arithimatical Crossover\\(AX)\cite{ref12} & $O = \alpha p^1 + (1 - \alpha) p^2$ & $\alpha \in (0,1)$\\
		Blend crossover\\($\mbox{BLX}-\alpha$)\cite{ref13} & $ \begin{aligned}O = \mathrm{uniform}\big(&p^1 - \alpha(p^2 - p^1), \\&p^2 + \alpha(p^2 - p^1)\big)\end{aligned} $ & $\alpha \in (0,1)$\\
		Simulated Binary Crossover\\(SBX)\cite{ref9} & \shortstack{$O^1= 0.5 \left[ (1 + \beta) p^1 + (1 - \beta) p^2 \right]$\\ $O^2= 0.5 \left[ (1 - \beta) p^1 + (1 + \beta) p^2 \right]$} &$\beta = 
		\begin{cases} 
		(2u)^{\frac{1}{\eta_c + 1}}, & \text{if } u \leq 0.5 \\
		\left( \frac{1}{2(1-u)} \right)^{\frac{1}{\eta_c + 1}}, & \text{if } u \geq 0.5 
		\end{cases}$ \\
		Laplace Crossover\\(LX)\cite{ref6} & \shortstack{$O^1 = p^1+\beta|p^1 - p^2|$\\$O^2 = p^2+\beta|p^1 - p^2|$} &$\beta=\begin{cases}a - b\log_{e}(u),& \text{if } u\leq0.5\\a + b\log_{e}(u),& \text{if } u > 0.5\end{cases}$\\
		Simplex crossover\\(SX)\cite{ref14} & $O=(1+\epsilon)\left(x^1 - \frac{1}{n}\sum_{i = 1}^{n}x^i\right)$ & $\begin{aligned}\varepsilon \geq 0\\~n \geq 3\end{aligned}$\\
		\hline
		\end{tabular}
	\end{table}

The relative crossover operator is presented in Table \ref{tab:linear}. These crossover operators solely employ the linear combination of parents to generate offspring. The exploration and convergence of the algorithm in the solution space are achieved by controlling the values of parameters. The differences among these crossover operators are mainly reflected in two aspects. One is the number of parents such as Simplex crossover(SX)[9] is multi-parents when others are two-parents, and the other is the way of determining the coefficient values like Laplace crossover uses Laplace distribution to generate coefficient and Logistic crossover uses Logistic distribution[6]. However, the offspring produced by these operators may not consistently evolve toward improved solutions due to the influence of low-fitness individuals in the population. This will lead to a slower convergence speed of the algorithm.

	\begin{table}
		\caption{Crossover operator with additional spatial information}\label{tab:sptial}
		\renewcommand{\arraystretch}{2.6}
		\begin{tabular}{ m{3cm} m{6cm} m{6cm} }
		\hline
		Name of crossover operation &  Formula & Other spatial information\\
		\hline
		PCX\cite{ref15} & $\vec{y} = \vec{x}^{(p)} + w_{\xi} \vec{d}^{(p)} + \sum_{i = 1, i \neq p}^{\mu} w_{\eta} \bar{D} e^{(i)}$ & \multirow{2}{*}{\parbox[t][6cm][t]{5cm}{Both the perpendicular distance from a point to a line and the spatial orthonormal basis are utilized to guide the direction of offspring generation.}}  \\
		UNDX\cite{ref16} & $\vec{y} = \vec{g} + \sum_{i = 1}^{\mu - 1} w_i \lvert \vec{d}^{(i)} \rvert e^{(i)} + \sum_{i = \mu}^{n} v_i D \vec{e}^{(i)}$ &\\
		\hline
		\end{tabular}
	
	\end{table}

The other crossover operators incorporate additional spatial information to generate offspring, typical related operators include Parent centric crossover(PCX) and Unimodal normally distributed crossover(UNDX), whose crossover formulas are presented in Table \ref{tab:sptial}. Although these operators perform outstandingly in some test problems, their high computational complexity leads to low algorithm operation efficiency, making them unsuitable for problems with high timeliness requirements.\\

\section{The proposed PSO-Inspired crossover operator}
Particle Swarm Optimization (PSO) is inspired by the collective behavior of biological swarms (e.g., bird flocks, fish schools) in adapting to environments, locating resources, and evading predators through information sharing. Hassan et al.\cite{ref17} have demonstrated through experiments that the convergence speed and computational efficiency of the PSO algorithm are significantly higher than those of the genetic algorithm. However, the mutation operator in the GA algorithm enables the algorithm to exhibit superior performance in escaping from local optima\cite{ref18}.

Therefore, extensive research has focused on integrating PSO and GA algorithms. Most of these hybrid approaches are based on the combination at the algorithm framework level. For example, the hybrid PSO - GA algorithm proposed in\cite{ref19,ref20} runs the PSO algorithm and the GA algorithm independently in two stages. The results of the one algorithm are used as the initial population for the another algorithm to perform further optimization. X.H. Shi et al. proposed a PSO-GA algorithm which allows the two algorithms run independently, and then, under the set number of iterations, partial high- quality populations are exchanged. R Huang et al. \cite{ref21} use PSO as the main framework and the population is divided into three groups during the iteration process, and applying GA-based crossover and mutation operations to each group separately, and finally the update formula of PSO itself is applied.

Traditional crossover operators only allow individuals in the current generation to perform crossover. The stochastic nature of selection may result in the inclusion of suboptimal individuals. As a result, Crossover may disrupt elite genes, causing offspring fitness divergence and superior gene loss, thereby impeding convergence. Inspired by the PSO algorithm, we proposes a novel PSO-Inspired crossover operator named PSOX. It mimics the learning method of the PSO algorithm and enables a more fine- grained integration of the PSO and GA algorithms, rather than a crude combination of frameworks. This crossover enables individuals to incorporate information from both the global best solution and other individuals' historical optima, so as to inherit excellent genes and accelerate the convergence process. The offspring are given by the equations:

	\begin{equation}
	O = w \cdot p_i + c_1 r_1 (pbest_j - p_i) + c_2 r_2 (gbest - p_i), \text{where } i \neq j
	\end{equation}

Whereare $w, c_1, c_2$ constant terms, which respectively determine the extent to which the offspring learn from themselves, the historical optimal solutions of other individuals, and the global historical optimal solution. Different from the setting in traditional crossover algorithms where both parents must come from the current generation, the PSOX algorithm allows individuals to perform inter-generational crossover to inherit the optimal genes of the global and other individuals. Meanwhile, random number $r_1, r_2$ withinare employed to prevent premature convergence to local optima.

\section{Experimental setup}
Three experiments were designed. Firstly, Experiment 1 (Section 4.1) evaluated the proposed crossover operator against benchmark operators using 15 test functions to assess algorithmic accuracy and stability. Secondly, Experiment 2(section 4.2) focused on convergence speed comparison: four test functions were selected from Experiment 1 results, on which PSOX performed relatively poorly while other operators showed good performance, to analyze convergence speed under the context of small population sizes. Thirdly, Experiment 3(section 4.3) explored the effects of different mutation rates on GAs equipped with PSOX.

\subsection{Accuracy and Stability Test}
To systematically evaluate the proposed algorithm's performance, fifteen 30-dimensional benchmark functions were employed to compare the novel crossover operator against five existing operators. These problems exhibit varying levels of difficulty and multi-modality characteristics. Meanwhile,To examine the crossover algorithm's sensitivity to mutation operators, all crossover variants were paired with both Gaussian mutation (GM) and non-uniform mutation (NUM) operators. The real-coded genetic algorithm (GA) incorporating the PSO-inspired crossover (PSOX) operator was simulated, with performance evaluated through mean and standard deviation metrics.

	\begin{table}
		\caption{Experimental parameter settings}\label{tab: experiment}
		\begin{tabular}{m{1.5cm} m{1.5cm} m{1.5cm} m{2cm} m{2cm} m{1.5cm} m{1.5cm} m{1.5cm} }
		\hline
		No.of turns & No.of generations & No.of Population & Population initial&Select operator&Crossover rate&Mutation operator&Mutation rate\\
		\hline
		30 & 1000 & 300 & Uniform random & Tournament with size=3 & 0.8& NUM & 0.1\\
		30 & 1000 & 300 & Uniform random & Tournament with size=3 & 0.8 & GM & 0.1\\
		\hline
		\end{tabular}
	\end{table}
	
	\begin{table}
		\caption{Parameter settings of the crossover operator}\label{tab: param}
		\begin{tabular}{ m{1.8cm} m{1.5cm} m{2.5cm} m{1.5cm} m{1.5cm} m{1.5cm} m{3cm} }
		\hline
		 & AX & FX &$\mbox{BLX}-\alpha$&SBX&LX&PSOX\\
		\hline
		Parameter1& $\alpha = 0.5$  & $\alpha = \mathrm{uniform}(0, 1)$  & $\alpha = 0.5$  &eta=2 &a=0 &w=0.6\\
		Parameter2&  &  &  & &b=0 &$c_1=1.5, c_2=1.5$\\
		\hline
		\end{tabular}
	\end{table}

Table \ref{tab: experiment} presents the uniform testing environment for all algorithms, with fixed parameter values maintained throughout every run. The parameter configurations for comparative crossover operators follow Table \ref{tab: param} specifications. All implementations utilize Python on a Windows platform with 2.9GHz processor and 32GB RAM.

\subsection{Convergence speed Test}
In Experiment 1, we only focused on the accuracy and stability of the results after a fixed number of iterations, without examining the performance during testing process. In Experiment 2, we selected four test functions (Test cases 4, 5, 7, 11) on which PSOX performs relatively poorly while other crossover operators show good performance. Under the conditions of Gaussian mutation(GM) as the mutation operator, a population size of 100, and 500 iterations, with all other settings remaining the same as in Experiment 1, thirty independent trials were performed to assess both convergence speed and process stability of the proposed algorithm.

\subsection{Effects of Mutation rate Test}
The experimental results from both Experiment 1 and Experiment 2 confirm PSOX's robust convergence characteristics and operational stability. However, considering that PSOX is inspired by the PSO algorithm, it may inherit the drawback of being prone to getting trapped in local optimal solutions. The mutation rate in genetic algorithms critically governs population diversity. By combining PSOX with different mutation rates, under the experimental conditions of Gaussian mutation(GM) as the mutation operator, a population size of 100, and 100 iterations, we conducted 30 independent experiments for each mutation rate parameter to analyze the impact of the mutation rate on the genetic algorithm equipped with PSOX.

\section{Results and discussion}
This section presents the analysis and summary of three experiments in Sections 6.1, 6.2, and 6.3, offering a comprehensive evaluation of the proposed PSOX operator from multiple perspectives.

\subsection{Performance Comparison of Different Crossover Operators}
The performance evaluation proceeds in two stages: (1) comparative analysis of means and standard deviations across all GAs to assess algorithmic performance and stability, followed by (2) application of the Kruskal-Wallis H-test ($\alpha = 0.05$) to detect significant inter-algorithm performance differences on identical test functions. A significant Kruskal-Wallis H-test result (+) indicates performance differences among algorithms, while a non-significant result ($\sim$) suggests no statistical difference.

Subsequently, Dunnett's test ($\alpha = 0.05$) is performed following a significant Kruskal-Wallis H-test result. In this test, the PSOX operator proposed in this paper serves as the control group, while other operators act as experimental groups for a one-sided great significance test. A significant Dunnett's test result (+) indicates superior performance in experimental groups compared to the control, while a non-significant result ($\sim$) shows no statistical difference.

\subsubsection{Performance Comparison of Different Operators Combined with the NUM Mutation Operator}

Table \ref{tab:NUM-results} presents the mean±SD of optimal objective values across generations for each test case, with corresponding Dunnett test p-values in Table \ref{tab:NUM-p}. Among the 15 test functions, the optimal values found by PSOX are not significantly smaller only in 4 test cases (Cases 3, 4, 14, and 15). In the remaining test cases, the optimal values obtained by PSOX are significantly smaller than those of other crossover operators. Notably, in Test Cases 1, 8, 9, 10, 11, and 13, the average optimal values of the fitness function found by PSOX are several orders of magnitude higher than those of other algorithms. Meanwhile, by comparing the standard deviations, the results of PSOX are slightly larger in Cases 5, 14, and 15, while in other cases, this value is far smaller than that of other crossover operators.

	\begin{landscape}
		\begin{table}[p]
		\centering
		\caption{Mean and Std of objective function values and Kruskal-Wallis H-test for mean objective function values for all GAs with NUM.}
		\label{tab:NUM-results}
		\small
		\renewcommand{\arraystretch}{1.2} 
		\begin{tabular}{c*{6}{cc}c}
		\toprule
		\multirow{2}{*}{\parbox{1cm}{Problem\\number}} & \multicolumn{2}{c}{AX-NUM} & \multicolumn{2}{c}{FX-NUM} & \multicolumn{2}{c}{BLX-$\alpha$-NUM} & \multicolumn{2}{c}{SBX-NUM} & \multicolumn{2}{c}{LX-NUM} & \multicolumn{2}{c}{PSOX-NUM} & \multirow{2}{*} {\parbox{1cm}{Kruskal-Wallis\\H-test}} \\
	    & Mean & std & Mean & std & Mean & std & Mean & std & Mean & std & Mean & std & \\
			\midrule
				1 & 1.7E+00 & 7.0E-01 & 7.8E-01 & 9.2E-01 & 2.7E-03 & 2.0E-03 & 2.2E+00 & 7.5E-01 & 6.5E-01 & 9.4E-01 & 6.8E-16 & 9.0E-16 & + \\
				2 & 2.9E-05 & 1.6E-05 & 2.1E-05 & 1.5E-05 & 1.4E-06 & 1.4E-06 & 1.9E-06 & 1.5E-06 & 6.3E-07 & 7.1E-07 & 4.3E-07 & 3.1E-07 & + \\
				3 & 9.3E-01 & 8.2E-02 & 4.7E-01 & 2.9E-01 & 1.5E-01 & 2.4E-01 & 5.4E-01 & 2.6E-01 & 7.3E-02 & 1.5E-01 & 0.0E+00 & 0.0E+00 & + \\
				4 & 2.2E-01 & 1.5E-01 & 1.8E-01 & 7.2E-02 & 2.2E-01 & 1.7E-01 & 2.3E-01 & 2.0E-01 & 1.3E-01 & 1.1E-01 & 3.5E-01 & 8.3E-02 & + \\
				5 & 2.4E-01 & 9.6E-02 & 1.0E-01 & 5.3E-02 & 3.2E-02 & 3.1E-02 & 1.9E-01 & 9.9E-02 & 2.6E-02 & 3.6E-02 & 9.6E-01 & 4.9E-01 & + \\
				6 & 1.2E+01 & 3.5E+00 & 1.5E+01 & 4.0E+00 & 1.9E+01 & 3.7E+00 & 2.0E+01 & 4.1E+00 & 2.0E+01 & 4.7E+00 & 3.2E+00 & 8.8E+00 & + \\
				7 & 9.0E+01 & 5.4E+01 & 6.2E+01 & 3.3E+01 & 1.1E+02 & 1.4E+02 & 2.2E+02 & 3.4E+02 & 8.1E+01 & 1.3E+02 & 2.8E+01 & 1.5E-01 & + \\
				8 & 8.3E+00 & 2.9E+00 & 3.5E+01 & 1.8E+01 & 5.1E+01 & 1.4E+01 & 6.7E+01 & 1.4E+01 & 2.8E+01 & 8.4E+00 & 0.0E+00 & 0.0E+00 & + \\
				9 & 8.5E-04 & 5.2E-04 & 1.6E-04 & 1.0E-04 & 9.7E-07 & 8.6E-07 & 5.3E-04 & 3.5E-04 & 8.3E-07 & 5.4E-07 & 0.0E+00 & 0.0E+00 & + \\
				10 & 1.0E-02 & 5.1E-03 & 1.8E-03 & 1.2E-03 & 1.0E-05 & 8.7E-06 & 5.7E-03 & 2.5E-03 & 1.0E-05 & 7.4E-06 & 0.0E+00 & 0.0E+00 & + \\
				11 & 2.2E+00 & 3.7E-01 & 3.1E+00 & 7.9E-01 & 6.2E+00 & 2.0E+00 & 8.8E+00 & 2.5E+00 & 6.9E+00 & 2.0E+00 & 3.9E-17 & 0.0E+00 & + \\
				12 & 9.3E-01 & 3.2E-01 & 9.1E-01 & 3.5E-01 & 3.1E-01 & 1.2E-01 & 1.1E+00 & 4.0E-01 & 4.8E-01 & 1.8E-01 & 1.1E-03 & 6.5E-04 & + \\
				13 & 1.9E+04 & 1.0E+04 & 3.5E+03 & 2.1E+03 & 1.4E+01 & 1.1E+01 & 1.1E+01 & 7.4E+03 & 1.5E+01 & 1.1E+01 & 0.0E+00 & 0.0E+00 & + \\
				14 & 2.2E-01 & 7.1E-02 & 1.7E-01 & 7.4E-02 & 1.6E-01 & 1.8E-01 & 2.5E-01 & 1.8E-01 & 1.2E-01 & 1.0E-01 & 3.4E-01 & 1.7E-01 & + \\
				15 & 2.2E-01 & 7.5E-02 & 1.0E-01 & 5.0E-02 & 2.8E-02 & 3.7E-02 & 1.5E-01 & 7.1E-02 & 1.8E-02 & 3.7E-02 & 1.1E+00 & 4.4E-01 & + \\
			\bottomrule
		\end{tabular}
		\end{table}
	\end{landscape}

	\begin{table}[htbp]
	    \centering
	    \caption{P - value and result of Dunnett Test for all GAs with NUM}
	    \begin{tabular}{cccccccccccc}
	        \toprule
	        \multirow{2}{*}{\shortstack{Problem\\number}} & \multicolumn{2}{c}{AX-NUM} & \multicolumn{2}{c}{FX-NUM} & \multicolumn{2}{c}{BLX-$\alpha$-NUM} & \multicolumn{2}{c}{SBX-NUM} & \multicolumn{2}{c}{LX-NUM} \\
	        \cmidrule(lr){2-3} \cmidrule(lr){4-5} \cmidrule(lr){6-7} \cmidrule(lr){8-9} \cmidrule(lr){10-11}
	         & P-value & result & Mean & std & Mean & std & Mean & std & Mean & std \\
	        \midrule
				1 & 1.0000 & + & 1.0000 & + & 0.8378 & + & 1.0000 & + & 1.0000 & + \\
				2 & 1.0000 & + & 1.0000 & + & 0.9296 & + & 0.9555 & + & 0.8572 & + \\
				3 & 1.0000 & + & 1.0000 & + & 1.0000 & + & 1.0000 & + & 0.9962 & + \\
				4 & 0.0008 & $\sim$ & 0.0000 & $\sim$ & 0.0009 & $\sim$ & 0.0037 & $\sim$ & 0.0000 & $\sim$ \\
				5 & 0.0000 & $\sim$ & 0.0000 & $\sim$ & 0.0000 & $\sim$ & 0.0000 & $\sim$ & 0.0000 & $\sim$ \\
				6 & 1.0000 & + & 1.0000 & + & 1.0000 & + & 1.0000 & + & 1.0000 & + \\
				7 & 0.9971 & + & 0.9739 & + & 0.9993 & + & 1.0000 & + & 0.9938 & + \\
				8 & 1.0000 & + & 1.0000 & + & 1.0000 & + & 1.0000 & + & 1.0000 & + \\
				9 & 1.0000 & + & 0.9999 & + & 0.8376 & + & 1.0000 & + & 0.8370 & + \\
				10 & 1.0000 & + & 1.0000 & + & 0.8383 & + & 1.0000 & + & 0.8383 & + \\
				11 & 1.0000 & + & 1.0000 & + & 1.0000 & + & 1.0000 & + & 1.0000 & + \\
				12 & 1.0000 & + & 1.0000 & + & 1.0000 & + & 1.0000 & + & 1.0000 & + \\
				13 & 1.0000 & + & 1.0000 & + & 0.8364 & + & 1.0000 & + & 0.8365 & + \\
				14 & 0.0014 & $\sim$ & 0.0000 & $\sim$ & 0.0000 & $\sim$ & 0.0268 & $\sim$ & 0.0000 & $\sim$ \\
				15 & 0.0000 & $\sim$ & 0.0000 & $\sim$ & 0.0000 & $\sim$ & 0.0000 & $\sim$ & 0.0000 & $\sim$ \\
	        \bottomrule
	    \end{tabular}
	    \label{tab:NUM-p}
	\end{table}

\subsubsection{Performance Comparison of Different Operators Combined with the GM Mutation Operator}

Based on the observations from Tables \ref{tab:GM-results} and \ref{tab:GM-p}, Test Case 2 exhibits consistent performance across all crossover operators. For Test Case 2, all GAs find same solution and Kruskal-Wallis H-test result shows negative. For Test Case 15, the PSOX operator performs worse than AX, FX, and BLX-. In all remaining test cases, it demonstrates superior performance. Similar to PSOX-GM-GA, It demonstrates superior performance in Test Cases 1, 3, 8, 9, 10, 11, and 13. Moreover, the algorithm shows excellent stability, successfully finding the solution in all 30 independent trials.

	\begin{landscape}
		\begin{table}[p]
		\centering
		\caption{Mean and Std of objective function values and Kruskal-Wallis H-test for mean objective function values for all GAs with NUM.}
		\label{tab:GM-results}
		\small
		\renewcommand{\arraystretch}{1.2} 
		\begin{tabular}{c*{6}{cc}c}
		\toprule
		\multirow{2}{*}{\parbox{1cm}{Problem\\number}} & \multicolumn{2}{c}{AX-GM} & \multicolumn{2}{c}{FX-GM} & \multicolumn{2}{c}{BLX-$\alpha$-GM} & \multicolumn{2}{c}{SBX-GM} & \multicolumn{2}{c}{LX-GM} & \multicolumn{2}{c}{PSOX-GM} & \multirow{2}{*} {\parbox{1cm}{Kruskal-Wallis\\H-test}} \\
	    & Mean & std & Mean & std & Mean & std & Mean & std & Mean & std & Mean & std & \\
			\midrule
			1 & 7.4E+00 & 1.1E+00 & 7.7E+00 & 1.2E+00 & 1.2E+01 & 2.3E+00 & 1.8E+01 & 4.1E-01 & 1.8E+01 & 1.0E+00 & 6.8E-16 & 9.0E-16 & + \\
			2 & 3.1E-07 & 0.0E+00 & 3.1E-07 & 0.0E+00 & 3.1E-07 & 0.0E+00 & 3.1E-07 & 0.0E+00 & 3.1E-07 & 0.0E+00 & 3.1E-07 & 0.0E+00 & $\sim$ \\
			3 & 1.4E+01 & 4.1E+00 & 1.2E+01 & 4.6E+00 & 3.7E+01 & 2.7E+01 & 2.2E+02 & 7.3E+01 & 2.0E+00 & 3.2E+00 & 0.0E+00 & 0.0E+00 & + \\
			4 & 1.4E-02 & 4.5E-02 & 1.0E-02 & 3.2E-02 & 4.4E-01 & 7.7E-01 & 1.1E+00 & 8.9E-01 & 3.8E-01 & 5.0E-01 & 5.4E-05 & 2.0E-05 & + \\
			5 & 1.1E-03 & 3.4E-03 & 8.6E-04 & 2.8E-03 & 2.1E-02 & 7.8E-02 & 4.6E-01 & 8.2E-01 & 2.5E-01 & 5.6E-01 & 9.6E-02 & 1.3E-01 & + \\
			6 & 3.5E+01 & 1.3E+01 & 4.9E+01 & 1.5E+01 & 1.2E+02 & 3.6E+01 & 1.7E+02 & 2.8E+01 & 1.7E+02 & 2.8E+01 & 8.8E+00 & 1.6E+01 & + \\
			7 & 1.0E+02 & 1.0E+02 & 1.3E+02 & 1.5E+02 & 2.2E+04 & 9.3E+04 & 3.7E+05 & 4.1E+05 & 2.8E+02 & 4.2E+02 & 2.8E+01 & 1.8E-01 & + \\
			8 & 2.4E-02 & 1.1E-02 & 1.4E+00 & 1.8E+00 & 1.5E+00 & 1.1E+00 & 3.4E+00 & 2.3E+00 & 8.7E-01 & 6.5E-01 & 3.6E-239 & 0.0E+00 & + \\
			9 & 6.8E-05 & 2.5E-05 & 7.0E-05 & 2.2E-05 & 1.1E-04 & 3.7E-05 & 2.9E-04 & 8.6E-05 & 1.6E-04 & 5.0E-05 & 1.6E-232 & 0.0E+00 & + \\
			10 & 9.6E-04 & 3.0E-04 & 1.3E-03 & 5.7E-04 & 2.1E-03 & 7.6E-04 & 4.8E-03 & 1.5E-03 & 2.3E-03 & 1.0E-03 & 3.0E-231 & 0.0E+00 & + \\
			11 & 1.2E+01 & 2.5E+00 & 1.9E+01 & 4.0E+00 & 6.4E+01 & 7.7E+00 & 6.5E+01 & 5.8E+00 & 6.6E+01 & 6.1E+00 & 1.4E-119 & 7.8E-119 & + \\
			12 & 2.9E+00 & 9.8E-01 & 2.3E+00 & 8.1E-01 & 9.2E-01 & 4.8E-01 & 6.8E+00 & 2.2E+00 & 7.6E-01 & 2.4E-01 & 1.7E-03 & 9.4E-04 & + \\
			13 & 8.6E+02 & 2.6E+02 & 8.4E+02 & 3.3E+02 & 1.3E+03 & 4.1E+02 & 7.4E+03 & 3.3E+03 & 1.7E+03 & 5.0E+02 & 8.8E-228 & 0.0E+00 & + \\
			14 & 6.9E-03 & 2.6E-02 & 1.4E-02 & 4.5E-02 & 3.3E-01 & 4.4E-01 & 1.2E+00 & 9.4E-01 & 3.3E-01 & 4.5E-01 & 5.8E-05 & 4.7E-05 & + \\
			15 & 4.3E-04 & 2.0E-03 & 7.6E-04 & 2.8E-03 & 6.0E-02 & 2.0E-01 & 3.8E-01 & 5.2E-01 & 3.7E-01 & 7.3E-01 & 1.8E-01 & 3.6E-01 & + \\
			\bottomrule
		\end{tabular}
		\end{table}
	\end{landscape}

	\begin{table}[htbp]
	    \centering
	    \caption{P - value and result of Dunnett Test for all GAs with GM}
	    \begin{tabular}{cccccccccccc}
	        \toprule
	        \multirow{2}{*}{\shortstack{Problem\\number}} & \multicolumn{2}{c}{AX-GM} & \multicolumn{2}{c}{FX-GM} & \multicolumn{2}{c}{BLX-GM} & \multicolumn{2}{c}{SBX-GM} & \multicolumn{2}{c}{LX-GM} \\
	        \cmidrule(lr){2-3} \cmidrule(lr){4-5} \cmidrule(lr){6-7} \cmidrule(lr){8-9} \cmidrule(lr){10-11}
	         & P-value & result & Mean & std & Mean & std & Mean & std & Mean & std \\
	        \midrule
				1 & 1.0000 & + & 1.0000 & + & 1.0000 & + & 1.0000 & + & 1.0000 & + \\
				2 & - & - & - & - & - & - & - & - & - & - \\
				3 & 0.9987 & + & 0.9959 & + & 1.0000 & + & 1.0000 & + & 0.8959 & + \\
				4 & 0.8619 & + & 0.8551 & + & 1.0000 & + & 1.0000 & + & 1.0000 & + \\
				5 & 0.4655 & + & 0.4643 & + & 0.5530 & + & 1.0000 & + & 0.9964 & + \\
				6 & 1.0000 & + & 1.0000 & + & 1.0000 & + & 1.0000 & + & 1.0000 & + \\
				7 & 0.8338 & + & 0.8340 & + & 0.9403 & + & 1.0000 & + & 0.8350 & + \\
				8 & 0.8537 & + & 1.0000 & + & 1.0000 & + & 1.0000 & + & 1.0000 & + \\
				9 & 1.0000 & + & 1.0000 & + & 1.0000 & + & 1.0000 & + & 1.0000 & + \\
				10 & 1.0000 & + & 1.0000 & + & 1.0000 & + & 1.0000 & + & 1.0000 & + \\
				11 & 1.0000 & + & 1.0000 & + & 1.0000 & + & 1.0000 & + & 1.0000 & + \\
				12 & 1.0000 & + & 1.0000 & + & 1.0000 & + & 1.0000 & + & 1.0000 & + \\
				13 & 0.9999 & + & 0.9999 & + & 1.0000 & + & 1.0000 & + & 1.0000 & + \\
				14 & 0.8499 & + & 0.8655 & + & 1.0000 & + & 1.0000 & + & 1.0000 & + \\
				15 & 0.0151 & $\sim$ & 0.0152 & $\sim$ & 0.0352 & $\sim$ & 0.9995 & + & 0.9992 & + \\
	        \bottomrule
	    \end{tabular}
	    \label{tab:GM-p}
	\end{table}
\newpage
\subsection{Analysis of the Convergence Speed}
\begin{figure}
	\includegraphics[width=\textwidth]{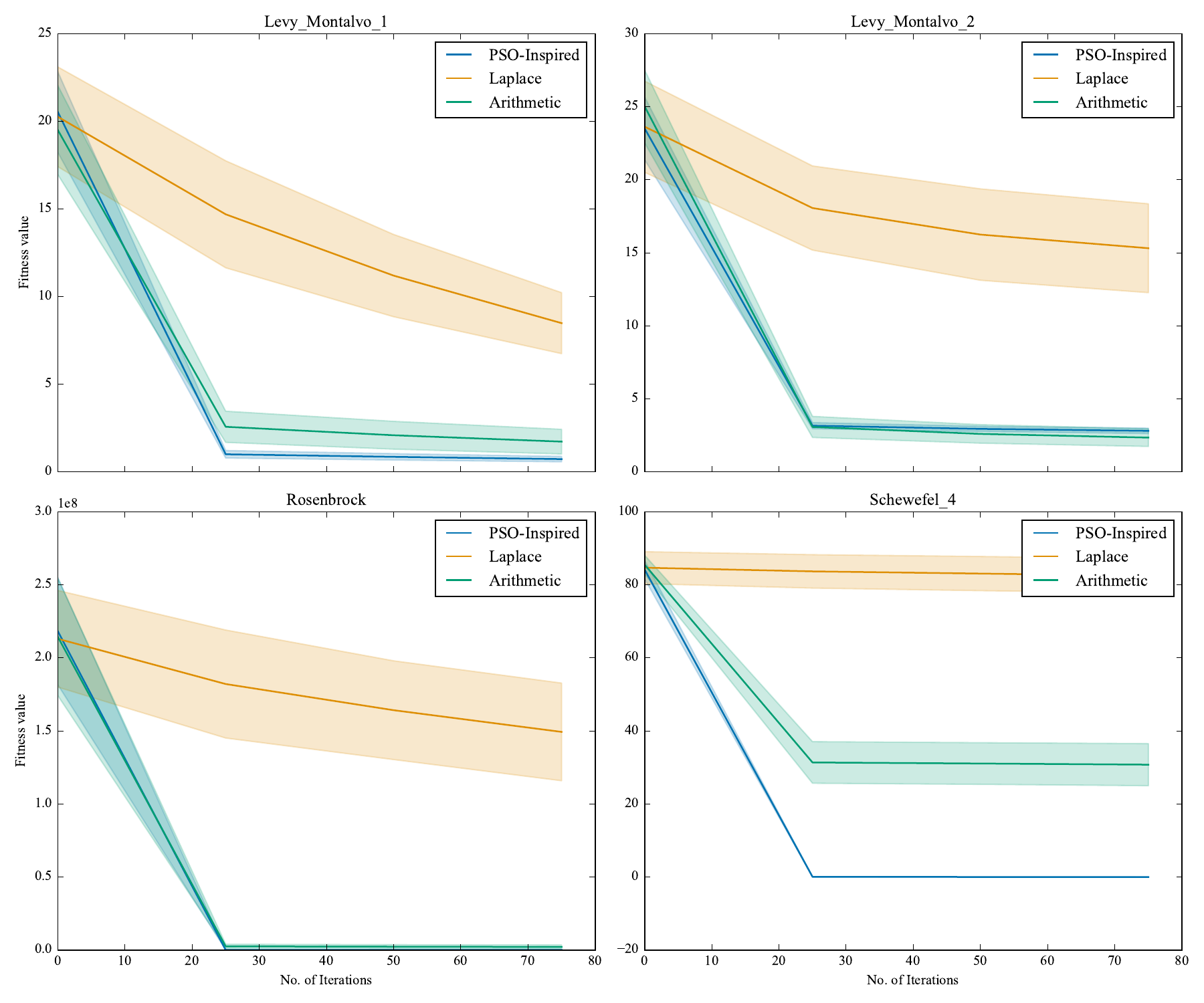}
	\caption{Comparison of the Convergence Speeds of Different Operators in Four Test Cases} \label{fig:speed}
	\end{figure}

For the test, the Laplace crossover (LX) operator and the Arithmetic crossover (AX) operator were selected as reference objects for comparison with the proposed PSOX operator. The comparison results are shown in Figure \ref{fig:speed}. The PSOX operator exhibits superior convergence speed compared to LX across all four test functions. For functions 5 (Levy Montalvo 2) and 7 (Rosenbrock), PSOX demonstrates comparable performance to AX. In function 4 (Levy Montalvo 1), PSOX achieves marginally faster convergence and significantly enhanced stability relative to AX. Furthermore, PSOX shows both accelerated convergence and improved stability in function 11 (Schewefel 4).
Overall, benefiting from its strategy of continuously learning from the globally optimal individual, the PSOX operator has remarkable convergence speed and convergence accuracy. Its characteristic of high stability ensures that the results obtained from a single run of the algorithm have a very high level of confidence, thus reducing the waste of resources caused by repeated testing.

\subsection{The Relationship between PSOX and the Mutation Rate}
\begin{figure}
	\includegraphics[width=\textwidth]{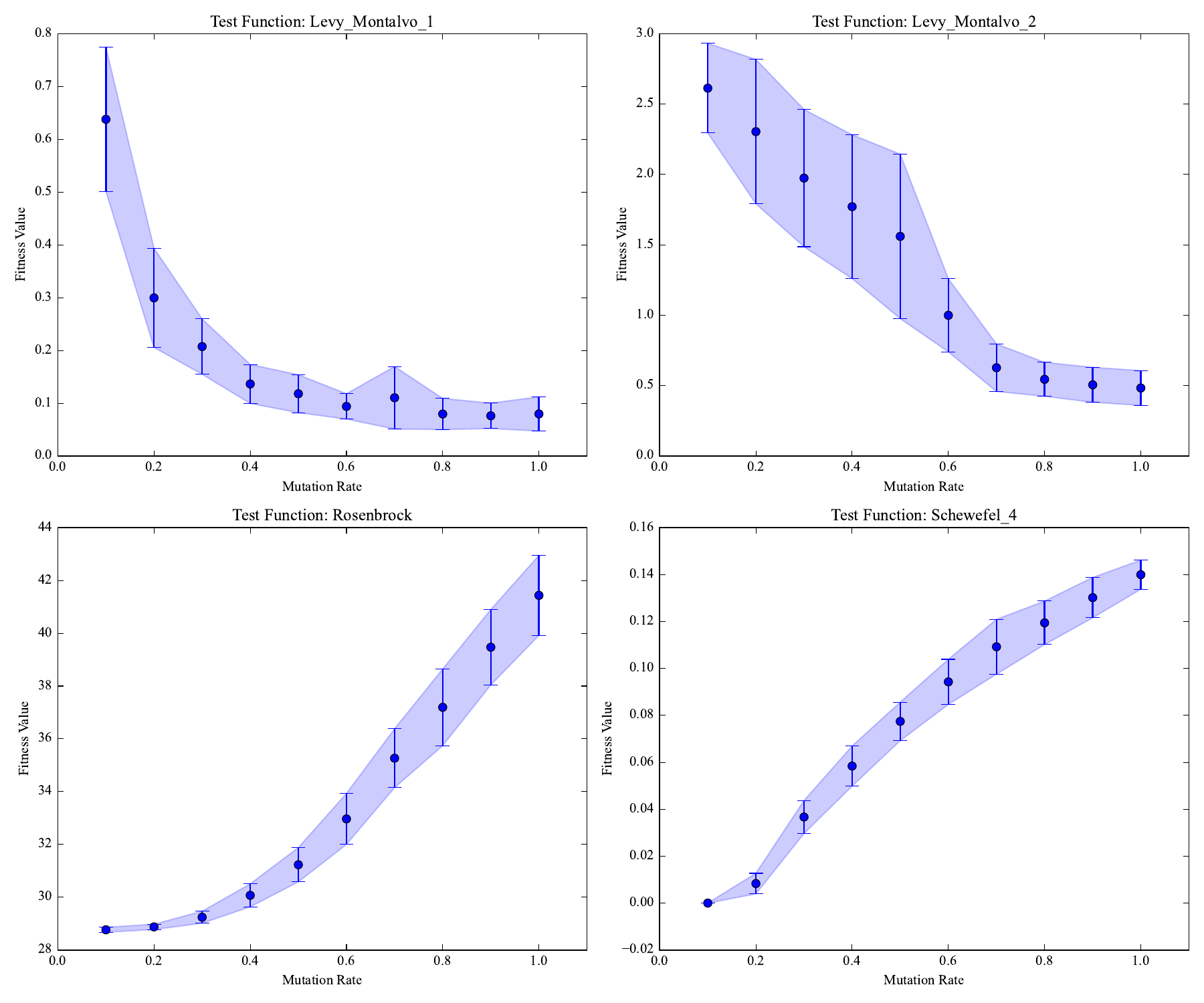}
	\caption{The Influence of the Mutation Rate on the GA with the PSOX Operator} \label{fig:mutation}
	\end{figure}

As clearly shown in Figure \ref{fig:mutation}, for Test Functions 4 and 5 (Levy Montalvo 1 and Levy Montalvo 2), the algorithm’s performance improves with the increase in mutation rate, and its stability gradually enhances. In contrast, for Test Functions 7 and 11 (Rosenbrock and Schewefel 4), the algorithm’s performance is inversely proportional to the mutation rate, and the standard deviation of the optimal fitness function values gradually increases.\\
Test Functions 4 and 5 are characterized by dense multi-peaks, complex oscillations, and multi-scale closed contour lines. These features make the PSOX operator primarily focused on convergence prone to falling into local optima. Therefore, a higher mutation rate is required to maintain population diversity and escape local optima. According to the operator’s formula, once an individual successfully escapes a local optimum, this information is rapidly disseminated to all individuals through crossover, enabling overall population evolution. In contrast, Test Functions 7 and 11 exhibit images with narrow valleys, separable chain structures, and strong regularity, which are more suitable for lower mutation rates. A higher mutation rate may disrupt important genes and lead to divergent results. 

\section{Conclusion}
This paper proposes a crossover operator for real - coded representation, inspired by the Particle Swarm Optimization (PSO) algorithm, called the PSO - Inspired Crossover operator (PSOX). PSOX is a multi - parent crossover method that allows cross - generational operation, simultaneously learning from the globally optimal individual and the historical optimal solutions of other individuals. Through three experiments, we comprehensively tested and analyzed the accuracy, stability, and convergence speed of the genetic algorithm configured with relevant crossover operators, and explored the relationship between the performance of PSOX and the mutation rate.

In the first experiment, we compared the PSOX operator with five other typical operators on 15 test functions. Regardless of whether the mutation operator was NUM or GM, in most test cases, PSOX outperformed other operators in terms of accuracy and stability. In a few test functions such as Function 4 and 5, due to the dense local optima and highly oscillatory terrain, PSOX showed poor performance. However, it was found in Experiment 3 that increasing the mutation rate could effectively solve such problems. Meanwhile, the performance of PSOX - GM was better than that of PSOX - NUM.

In the second experiment, four test functions were selected to compare PSOX with two other operators. It was found that the PSOX operator had a very fast convergence ability and stable convergence performance. The third experiment demonstrated the influence of different mutation operators on PSOX and provided suggestions for the application of PSOX: For problems with smooth, continuous, and highly regular solution spaces, even if there are many local optima, a lower mutation rate is more appropriate. For problems with high oscillations and multi - scale closed contours, a larger mutation rate should be used, and it can even be set to 1. Such a setting can significantly improve the solution results.

%
%
%
\bibliographystyle{unsrt}
\bibliography{Accelerating_Evolution_Integrating_PSO_Principles_into_Real-Coded_Genetic_Algorithm_Crossover}

\section*{Appendix}
1.Ackley's Problem
\begin{align*}
&\min_{x}f(x)= - 20\exp\left(-0.2\sqrt{\frac{1}{n}\sum_{i = 1}^{n}x_{i}^{2}}\right)-\exp\left(\frac{1}{n}\sum_{i = 1}^{n}\cos(2\pi x_{i})\right)+20 + e\\
&-30\leq x_{i}\leq30, x^{*}=(0,0,\ldots,0)\text{ and }f(x^{*}) = 0
\end{align*}
\\
2.Exponential Problem
\begin{align*}
&\min_{x}f(x)=-\exp\left(-0.5\sum_{i = 1}^{n}x_{i}^{2}\right)\\
&- 1\leq x_{i}\leq1,x^{*}=(0,\ldots,0)\text{ and }f(x^{*})=-1
\end{align*}
\\
3.Griewank Problem
\begin{align*}
&\min_{x}f(x)=1+\frac{1}{4000}\sum_{i = 1}^{n}x_{i}^{2}-\prod_{i = 1}^{n}\cos\left(\frac{x_{i}}{\sqrt{i}}\right)\\
&-600\leq x_{i}\leq600,x^{*}=(0,0,\ldots,0)\text{ and }f(x^{*}) = 0
\end{align*}
\\
4.Levy and Montalvo Problem 1
\begin{align*}
&\min_{x}f(x)=\left(\frac{\pi}{n}\right)\left(10\sin^{2}(\pi y_{1})+\sum_{i = 1}^{n - 1}(y_{i}-1)^{2}[1 + 10\sin^{2}(\pi y_{i + 1})]+(y_{n}-1)^{2}\right)\\
&\text{where }y_{i}=1+\frac{1}{4}(x_{i}+1),-10\leq x_{i}\leq10,x^{*}=(0,0,\ldots,0)\text{ and }f(x^{*}) = 0
\end{align*}
\\
5.Levy and Montalvo Problem 2
\begin{align*}
&\min_{x}f(x)=0.1\sin^{2}(3\pi x_{1})+\sum_{i = 1}^{n - 1}(x_{i}-1)^{2}(1 + \sin^{2}(3\pi x_{i + 1}))+(x_{n}-1)^{2}(1+\sin^{2}(2\pi x_{n}))\\
&-5\leq x_{i}\leq5,x^{*}=(0,0,\ldots,0)\text{ and }f(x^{*}) = 0
\end{align*}
\\
6.Rastrigin Problem
\begin{align*}
&\min_{x}f(x)=10n+\sum_{i = 1}^{n}(x_{i}^{2}-10\cos(2\pi x_{i}))\\
&-5.12\leq x_{i}\leq5.12,x^{*}=(0,0,\ldots,0)\text{ and }f(x^{*}) = 0
\end{align*}
\\
7.Rosenbrock Problem
\begin{align*}
&\min_{x}f(x)=\sum_{i = 1}^{n - 1}[100(x_{i + 1}-x_{i}^{2})^{2}+(x_{i}-1)^{2}]\\
&-30\leq x_{i}\leq30,x^{*}=(0,0,\ldots,0)\text{ and }f(x^{*}) = 0
\end{align*}
\\
8.Zakharov’s Function
\begin{align*}
&\min_{x}f(x)=\sum_{i = 1}^{n}x_{i}^{2}+\left(\sum_{i = 1}^{n}\frac{i}{2}x_{i}\right)^{2}+\left(\sum_{i = 1}^{n}\frac{i}{2}x_{i}\right)^{4}\\
&-5.12\leq x_{i}\leq5.12,x^{*}=(0,0,\ldots,0)\text{ and }f(x^{*}) = 0
\end{align*}
\\
9.Sphere Function
\begin{align*}
&\min_{x}f(x)=\sum_{i = 1}^{n}x_{i}^{2}\\
&-5.12\leq x_{i}\leq5.12,x^{*}=(0,0,\ldots,0)\text{ and }f(x^{*}) = 0
\end{align*}
\\
10.Axis Parallel Hyper Ellipsoid
\begin{align*}
\min_{x}f(x)=\sum_{i = 1}^{n}ix_{i}^{2}, -5.12\leq x_{i}\leq5.12, x^{*}=(0,0,\ldots,0)\text{ and }f(x^{*}) = 0
\end{align*}
\\
11.Schwefel Problem 4
\begin{align*}
&\min_{x}f(x)=\max_{i}|x_{i}|\\
&-100\leq x_{i}\leq100,x^{*}=(0,0,\ldots,0)\text{ and }f(x^{*}) = 0
\end{align*}
\\
12.De - Jong’s Function with Noise
\begin{align*}
&\min_{x}f(x)=\sum_{i = 1}^{n}(x_{i}^{4}+\text{rand}(0,1))\\
&-10\leq x_{i}\leq10,x^{*}=(0,0,\ldots,0)\text{ and }f(x^{*}) = 0
\end{align*}
\\
13.Cigar Function
\begin{align*}
&\min_{x}f(x)=x_{1}^{2}+10000000\sum_{i = 2}^{n}x_{i}^{2}\\
&-10\leq x_{i}\leq10,x^{*}=(0,0,\ldots,0)\text{ and }f(x^{*}) = 0
\end{align*}
\\
14.Generalized Penalized Function 1
\begin{align*}
&\min_{x}f(x)=\frac{\pi}{n}\left(10\sin^{2}(\pi y_{1})+\sum_{i = 1}^{n - 1}(y_{i}-1)^{2}[1 + 10\sin^{2}(\pi y_{i + 1})]+(y_{n}-1)^{2}\right)+\sum_{i = 1}^{n}u(x_{i},10,100,4)\\
&\text{where }y_{i}=1+\frac{1}{4}(x_{i}+1),-10\leq x_{i}\leq10,x^{*}=(0,0,\ldots,0)\text{ and }f(x^{*}) = 0
\end{align*}
\\
15.Generalized Penalized Function 2
\begin{align*}
&\min_{x}f(x)=0.1\sin^{2}(3\pi x_{1})+\sum_{i = 1}^{n - 1}(x_{i}-1)^{2}(1 + \sin^{2}(3\pi x_{i + 1}))+(x_{n}-1)^{2}(1+\sin^{2}(2\pi x_{n}))\\
&+\sum_{i = 1}^{n}u(x_{i},10,100,4),-5.12\leq x_{i}\leq5.12,x^{*}=(0,0,\ldots,0)\text{ and }f(x^{*}) = 0
\end{align*}
\\
In problem number 14 and 15, the value of penalty function u is given by the following expression:
\begin{equation}
u(x,a,k,m)=
\begin{cases}
k\cdot(x - a)^{m}&\text{if }x > a\\
-k\cdot(a - x)^{m}&\text{if }x < -a\\
0&\text{otherwise}
\end{cases}
\end{equation}

\end{document}